\documentclass[runningheads]{svproc}
\usepackage{graphicx}
\begin{document}
\title{Image Aesthetics Assessment using Multi Channel Convolutional Neural Networks}
\titlerunning{Image Aesthetics Assessment using Multi Channel CNNs}
%
\author{Nishi Doshi\inst{1} \and
Gitam Shikkenawis\inst{2} \and
Suman K Mitra\inst{1}}
\authorrunning{N. Doshi et al.}
%
\institute{Dhirubhai Ambani Institute of Information and Communication Technology, Gandhinagar, India  \\
\email{\{201601408,suman\_mitra\}@daiict.ac.in}\\
 \and
C R Rao Advanced Institute of Mathematics, Statistics and Computer Science, Hyderabad, India \\ 
\email{gitam365@gmail.com}}

\maketitle              
\begin{abstract}
Image Aesthetics Assessment is one of the emerging domains in research. The domain deals with classification of images  into categories depending on the basis of how pleasant they are for the users to watch. In this article, the focus is on categorizing the images in high quality and low quality image. Deep convolutional neural networks are used to classify the images. Instead of using just the raw image as input, different crops and saliency maps of the images are also used, as input to the proposed multi channel CNN architecture. The experiments reported on widely used AVA database show improvement in the aesthetic assessment performance over existing approaches.

\end{abstract}
\section{Introduction}
Image aesthetics is one of the emerging domains of research. Image Aesthetics Assessment (IAA) problem deals with giving rating to images on the basis of how pleasant they are for the user to watch. A human is more likely to feel happy looking at a high quality picture rather than low quality images. On the basis of aesthetic value of an image, it helps the user to identify whether he or she is more likely to like the image and view the rest or not. It deals with finding out the aesthetic quality of an image that is classifying an image as into the category of either high quality image or low quality image. 

\begin{figure}[!h]
    \centering
    \includegraphics[height=3.5cm,width=4cm]{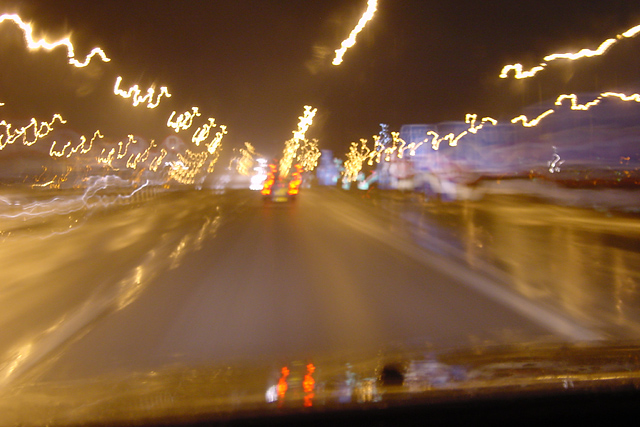}
    \includegraphics[height=3.5cm,width=4cm]{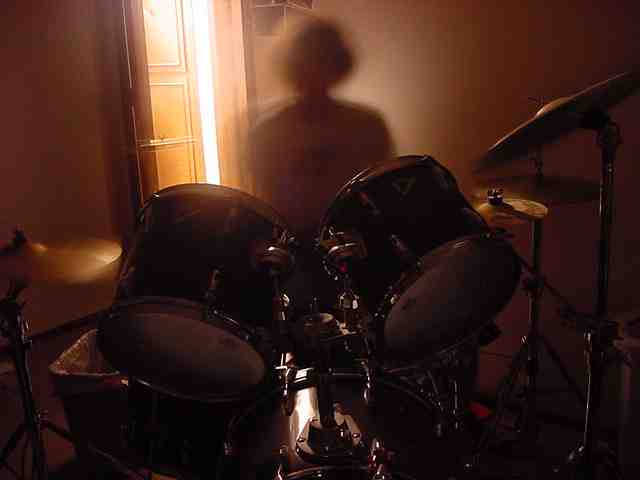}
    \includegraphics[height=3.5cm,width=4cm]{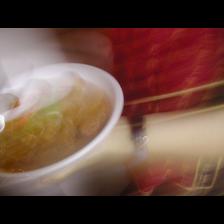}
    \caption{Low quality images}
    \label{fig:low}
\end{figure}

\begin{figure}[!h]
    \centering
    \includegraphics[height=3.5cm,width=4cm]{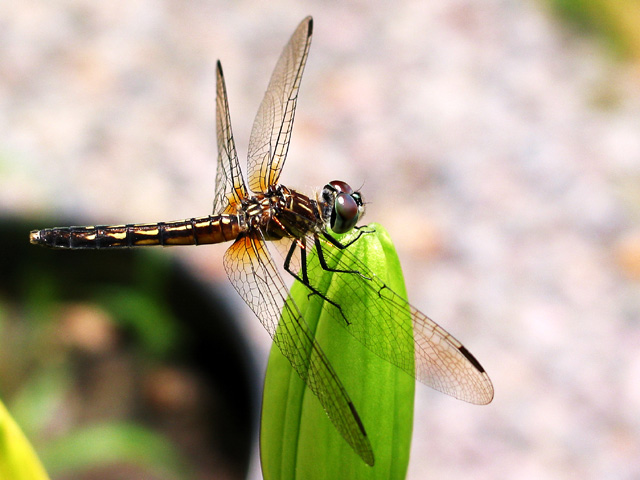}
    \includegraphics[height=3.5cm,width=4cm]{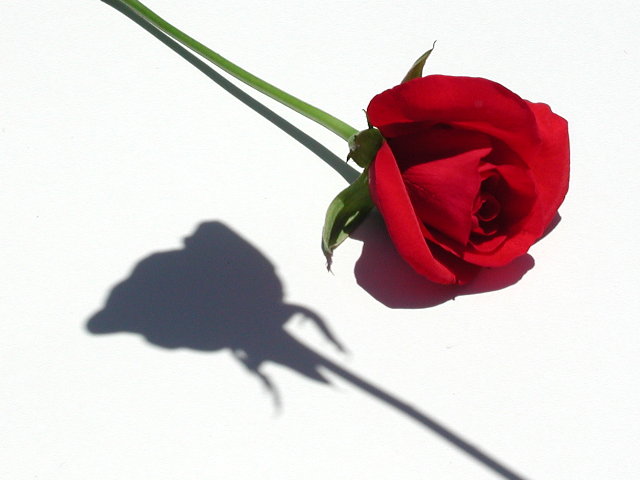}
    \includegraphics[height=3.5cm,width=4cm]{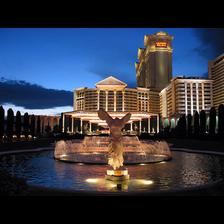}
    \caption{High quality images}
    \label{fig:high}
\end{figure}

A photo can be clicked from any device. Every device has certain resolution which results in the clarity in the pixels. Thus, use of different devices for clicking the photographs leads to the existence of this problem domain and its classification as well. It also depends on of the photo has been captured by the photographer and the scene covered. To demonstrate this, a few images from the AVA database \cite{one} are shown in Figure \ref{fig:low} and Figure \ref{fig:high}. 

The aesthetics assessment problem has been used in many practical applications specially to attract the users by showing visually more appealing images. Various search engines makes use of such a classification. For a given search word, it takes the aesthetic value of an image into consideration while showing any image on top or showing it last. Review applications, which allow the users to upload photographs and write review about the products/place also take aesthetics of the uploaded images into consideration. While displaying the review, photos with high aesthetics quality or clicked from a high quality camera are more likely to be shown first compared to a photo clicked from a low quality camera or having low aesthetic quality. 

Being one of the recent problems, a lot of research work is going in the domain of Image Aesthetics. Not only restricting the problem definition to classification; many variants of the problem statements exist and researchers are working on the same. RAPID : Rating Pictorial Aesthetics using Deep Learning was one of the earliest models used for classification  \cite{six}. It was the model to propose use of AlexNet architecture as well as double column architecture involving use of global and local views of an image as an input to channels. In Brain Inspired Deep Neural Network model \cite{seven}, 14 style CNNs are pre-trained, and they are parallelly cascaded and used as the input to a final CNN for rating aesthetic prediction, where the aesthetic quality score of an image is subsequently inferred on AVA dataset \cite{one}. Category Specific CNN having initially classification of images into different classes and then corresponding to every class a different CNN architecture is trained to classify images into different categories : either high quality or low quality \cite{two}. 

In this article, variants of multi column Convolutional Neural Networks (CNN) are proposed. In particular, the novelty of the approach lies in using informative inputs to the channels in multi column architecture. A pre-trianed CNN architecture - VGG19 \cite{VGG19} on ImageNet database  \cite{imagenet} is fine tuned for image aesthetics assessment. The images are classified into one of the two clasess, i.e. high aesthetic quality or low aesthetic quality. The experiments are performed on the widely used AVA databse \cite{three}.

The organization of the paper is as follows: Section \ref{sec:dl} discusses deep CNN architectures and fine tuning the pre-trained deep networks. Variants of multi channel CNN for IAA are discussed in Section \ref{sec:MCNN}. Experiments on AVA database are reported in Section \ref{sec:exp} followed by conclusions in \ref{sec:con}.

\section{Deep learning for IAA} \label{sec:dl}

In this paper, deep convolutional neural networks are used for assessing the aesthetic quality of the images. In particular, widely used deep learning architectures namely Alexnet \cite{four} and VGG19 \cite{VGG19} are used. The architectural details of both networks are discussed in this section along with the procedure of fine tuning an already trained deep neural network.

\subsection{CNN Architectures}

Mainly two CNN column architectures were involved in the models designed to classify the images : Alexnet and VGG19

\subsubsection{AlexNet}

\begin{figure}[!h]
    \centering
    \includegraphics[width=10cm,height=5cm]{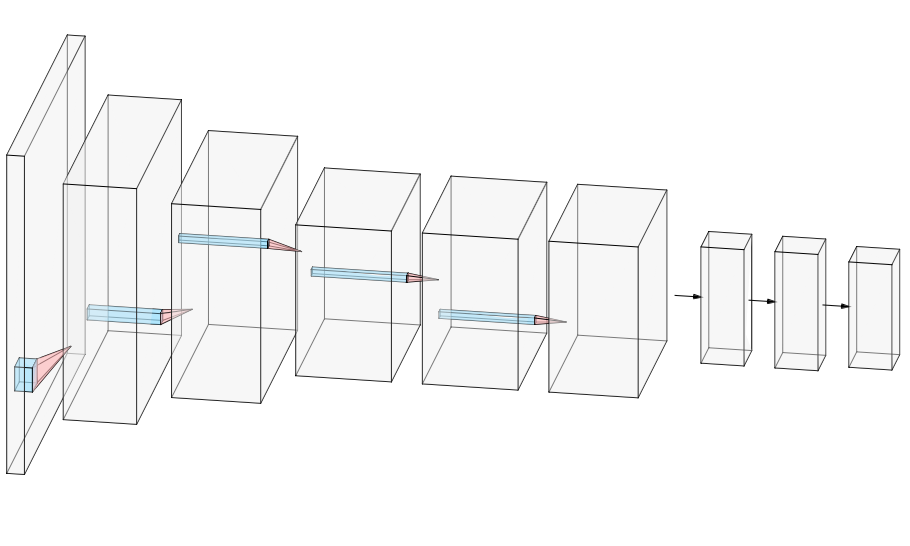}
    \caption{AlexNet Convolutional Neural Network architecture}
    \label{fig:figure2}
\end{figure}

AlexNet architecture involves a total of $8$ layers : $5$ convolutional layers and $3$ fully connected layers.  After every layer there is a ReLu function calculation. RAPID \cite{six} was the first method that used AlexNet architecture for solving the classification problem for image aesthetics. The architecture can be visualized by the Layer diagram in Figure \ref{fig:figure2}.

\subsubsection{VGG19}
  
VGG19 is a convolutional neural network architecture that is already trained on more than hundred million images of ImageNet database \cite{imagenet}. The network is $19$ layers deep and classifies images into $1000$ classes : some namely mug, mouse, keyboard, pencil. In considering the architecture for training and classifying images into two classes : High quality and Low quality images the initial network architecture is kept the same and the fully connected layers are changed to obtain classification into $2$ classes by adding $9$ dense layers having ReLu activation after Max Pooling layer of block $5$.

The VGG19 architecture is divided into $5$ blocks. Each block has certain number of convolutional layers with each block ending with Max Pooling layer. The number of convolutional layers per block are described in the Table \ref{tab:table6}.
\begin{table}[!h]
\begin{center}
    \begin{tabular}{|c|c|c|c|c|c|}
        \hline
         Block Number & $1$ & $2$ & $3$ & $4$ & $5$ \\
         \hline
         Number of convolutional layers & $2$ & $2$ & $4$ & $4$ & $4$\\
         \hline
    \end{tabular}
    \caption{Number of convolutional layers in different blocks of CNN}
    \label{tab:table6}
\end{center}
\end{table}

\subsection{Transfer Learning and Fine Tuning}
As discussed in CNN architectures, the Deep networks have huge number of parameters and training them requires very large databases. If large enough datasets are not available, transfer learning is shown to be working very well. In transfer learning, a model trained on a large and general dataset is used as the base generic model for new application. Thus, the learned features maps are used without having to start from scratch training a large model on a large dataset. In transfer learning, the weights of the convolution layers are used as it is, only the soft max layers are re-trianed to give desired output.  

In case of fine tuning, instead of using fixed weights, top  convolutional layers are made traininable. Along with fully connected soft max layers, some of the initial weights of convolutional layers (the layers near the fully connected layers that is the last last of the layers of the network) are also updated. 

In the current proposal, instead of training the deep CNN from scratch, pre-trained VGG19 network on ImageNet database is used as the base network. Top CNN layers and fully connected layers are re-trained to solve the problem of classification of images on the basis of their aesthetic value. 

\section{Multi Channel CNN based Image Aesthetic Assessment} \label{sec:MCNN}

The conventional methods of fine tuning a pre-trained deep network involve giving the original image as input to the single channel network and predict the high/low aesthetic quality. However, as suggested in \cite{six}, instead of taking just the input image for assessing the quality, considering global and local views of the image in a double column network enhances the classification performance of the IAA technique. In this section, various image pre-processing and feature selection procedures are discussed. Based on the observations made, a double and triple column architecture for IAA is proposed. 

\subsection{Image Pre-processing and Feature Selection}

As already designed deep CNN architectures are used in this article, the sizes of the images to be given as input are fixed, hence the images are required to be resized according to the specifications of respective networks. Both Alexnet and VGG19 support input sizes of $224 \times 224$ whereas AVA dataset contains images with different sizes. Hence, all the images were resized to $224 \times 224$. In addition to resizing the original image, input images by padding and cropping various portions are generated to negate the effects of scaling and resizing. Also, instead of using the raw image as a input, saliency map are used as input to one of the channels. These pre-processing and feature selection techniques are discussed in the following section.

\subsubsection{Original Image}
  
As both VGG19 and AlexNet architectures require images of size $224\times224$ as input, the images of the dataset were resized to the desired input size. As the pixels become blur after resizing, the aspect ratio of image was taken into consideration while resizing.
  
\subsubsection{Padded Images}
  
To negate the impact of aspect ratio, image was initially padded with zeros to make it a square image. And then the image was resized to $224\times224$. For example an image of $512\times1024$ dimensions  was first converted to image of size $1024\times1024$ by padding the image on top and bottom with $256$ zeros, and then this $1024\times1024$ was resized to $512\times512$ image.
  
\subsubsection{Cropped portion of Image}
  
Human eyes generally do not look at the entire image as a whole. In computational aspects, crops of image can be used to detect such portions which can be focus of human eye and be responsible for the judgement about the image. In this paper, different cropping techniques are used to capture the essence of the image instead of resizing it to fit the input size. 
\\
\textbf{\textit{Center Crop}}
\\
Center portion of the image is cropped and given as input to channel \cite{six}. The reason to take center crop of $224\times224$ size is that when a person looks at the picture; it is the center most part of the image where eyes of a human rest first. Human eye tends to see the center position of the image most rather than being focused on corner parts of an image. Hence, considering the center crop of the image would lead to decide relevant features for channel and help towards a better classification into two classes.
\\
\textbf{\textit{Random Crop}}
\\
Instead of using the fixed center crop, three random cropped patches of size $224\times224$ were generated for images. A novelty added to this approach was refining the algorithm to generate random crops. While generating these three crops, two things were taken into consideration: (1) it is not the center crop of the image and (2) it does not overlap with other random crops already taken. If the random cropped patches belonged to similar portion of the image, misleading results may be produced as same images are passed in both channels of the network. Hence, the current proposal takes into consideration that when generating random cropped patches of $224\times224$, the distance between the center of crops has to be more than a threshold, so that two crops belonging to same region or image are not generated. A distance of $100$ was kept for $x$ and $y$ coordinate of the generated crop image and previously generated crops of the  image.
  
\subsubsection{Saliency Map of Image}
While viewing an image, humans do not treat the entire scene equally, mostly the focus is on visually appealing parts. Saliency deals with unique features of image related to visual representation of an image. The saliency map highlights the pixels which have more visual importance in the image. It elaborates the part of image to which our brain gets attracted the most.

Here, we have considered use of static saliency map detection as images are static in nature. Two types of static saliency maps are taken into consideration.
\\
\\
\textbf{\textit{Spectral Residual Map}}\\
The algorithm analyzes the log-spectrum of an image, extracts the spectral residual and proposes a fast method to construct spectral residual saliency map which suggests the positions of visually attracted spots of an image.
\\
\textbf{\textit{Fine grained Map}}
\\
Human eyes have retina which consists of two types of cells : off center and on center. Specialty of these two types of cells are as follows :

\begin{itemize}
    \item On center : It responds to bright areas surrounded by dark background. 
    \item Off center : It responds to dark areas surrounded by bright background.
\end{itemize}
  
Fine grained saliency map is generated by taking considering the on center and off center differences \cite{Wang2014}. 

\begin{figure}[!h]
    \centering
    \includegraphics[width=4cm,height=4cm]{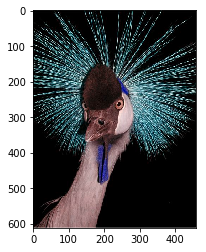}
    \includegraphics[width=4cm,height=4cm]{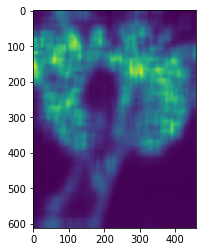}
    \includegraphics[width=4cm,height=4cm]{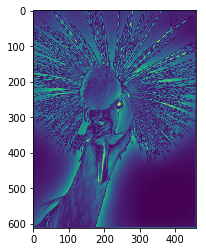}
    \caption{In order from left : Original high quality image taken from PhotoQuality Dataset \cite{five}, corresponding spectral residual saliency map and fine grained saliency map.}
    \label{fig:saliency}
\end{figure}

Spectral residual and fine grained saliency maps have been shown in Figure \ref{fig:saliency}. 

Thus, various ways of pre-processing and feature extraction from the original input image have been discussed in this Section. Next part discusses using these processed images as input to multi channel CNNs.  

\subsection{Proposed multi channel CNN for IAA }
Instead of working on a single channel with the raw image as input, in this article, we have used pre-processed images as discussed before in this Section as input along with the raw image. Thus, building a multi channel convolutional neural network architecture. Experiments have been performed on doble and triple column networks. The configuration details of both the networks are discussed below: 

\subsubsection{Double Column Network}
  
The double column network involves use of two pipelines and concatenation of those channels to generate output classifier. In one part of the network : original form of the image, the padded form of the image and the center cropped form of an image are supplied whereas in the second part of network three variants of random cropped forms of images are given. \\

In case of AlexNet architecture; initially trained single column network is used as base model for both channels. The parameters after concatenating are trained for $300$ epochs and then fine tuning is carried out by making the $4^{th}$ and $5^{th}$ convolutional layers trainable.  \\

In case of VGG network; a pre-trained VGG19 network on ImageNet dataset is taken as the base network for both the channels. As in cale of AlexNet, $4^{th}$ and $5^{th}$ convolutional layers are fine tuned for image aesthetics assessment task.

\subsubsection{Triple Column Network}
  
The triple column network involves use of three pipelines and concatenation of those channels to generate output classifier. There is an addition of third pipeline to double column network where two variants of saliency maps that is spectral residual map and fine grained maps are passed. \\ 

\begin{figure}[!h]
    \centering
    \includegraphics[width=10cm,height=5cm]{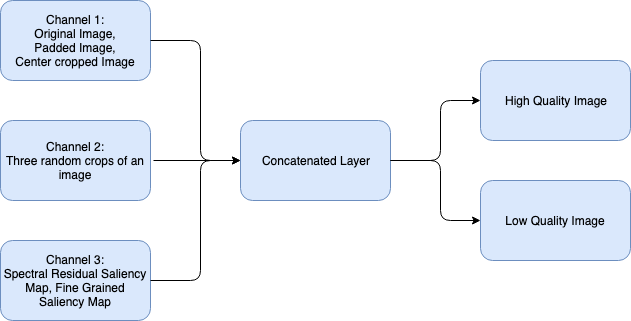}
    \caption{Triple column network design}
    \label{fig:figure1}
\end{figure}

VGG19 network is only trained for this column network. VGG19 network which had two channels already trained for double column network were used as base model and for third channel; single column network model was used as base model. After concatenating the results of all the three channels the classification of images into high quality and low quality was done. The network design for triple column network is shown in Figure \ref{fig:figure1}.

\section{Experiments} \label{sec:exp}

There are many datasets available for testing the validity of the model designed to solve the image aesthetic assessment problem. In this paper, the experimental results are reported on one of the most used database for IAA \textit{i.e.} AVA dataset \cite{three}.
 
\subsubsection{AVA Dataset}
 
AVA Dataset\cite{three} consists of images which have votes of users for every rating 1 to 10. As the model developed is for classifying images into two categories : high and low. The ratings were assigned to images on the basis of maximum number of votes corresponding to that image. The number of images corresponding to each rating are as follows :

\begin{table}[!h]
\begin{center}
\caption{ Number of images in AVA dataset corresponding to each and every vote}\label{tab:table3}
\begin{tabular}{|c|c|c|c|c|c|c|c|c|c|c|c|c|}
    \hline
     Rating & $1$ & $2$ & $3$ & $4$ & $5$ & $6$ & $7$ & $8$ & $9$ & $10$ & Total images\\
     \hline
     Number of images & $566$ & 104 & 1083 & $24305$ & $147483$ & $74294$ & $6824$ & $743$ & $31$ & $97$ & $255530$ \\
    \hline
\end{tabular}
\end{center}
\end{table}

As it can be seen from Table \ref{tab:table3}, out of $255530$ images, $147483$ that is around $58\%$ images belong to rating $5$. Hence, the images belonging to rating $1$, $2$, $3$ and $4$ were considered low quality images and images with rating $7$, $8$, $9$ and $10$ were considered high quality images.

\subsection{Comparison with various approaches}

Experiments using single column VGG19 and AlexNet architectures were carried out on AVA dataset. In case of AlexNet, the network is trained from scratch for image aesthetic assessment whereas for VGG19, a pre-trained model on ImageNet dataset is used. Here, the pre-trianed works as the base model and it is fine tuned for aesthetics assessment on AVA database. The results in terms of accuracy are reported in Table \ref{tab:table2}. It can be observed that there is significant improvement in testing accuracy in case of VGG19 as compared to AlexNet. Hence, the double and triple column CNN experiments have been performed by using the VGG19 architecture as the base CNN architecture. 

The results obtained after using double and triple channel CNNs are reported in Table \ref{tab:table2}. It can be observed that more that more $7\%$ enhancement in the testing accuracy is obtained using triple channel CNN over the double channel CNN. It shows that addition of saliency map as feature boost the IAA performance. 

For fair comparison, results on AVA dataset using a few existing deep neural network based approaches namely SCNN \cite{six}, DCNN \cite{six} and BDN \cite{seven} have also been reported in Table \ref{tab:table2}. It can be observed that the proposed triple channel architecture surpasses all three compared approaches.

\begin{table}[!h]
\caption{Different Architectures and Network results on AVA Dataset}\label{tab:table2}
\begin{center}
\begin{tabular}{|c|c|c|c|}
    \hline
     Architecture & Network & Train accuracy & Test Accuracy \\
     \hline
     AlexNet & Single Column &  $0.993$ & $0.6164$ \\
     \hline
     VGG19 
     
     & Single Column Network  & $0.9987$ & $0.7137$\\
     \hline
     VGG19 & Double Column Network & $0.8082$ & $0.7444$ \\
     \hline

     VGG19 & Triple Column Network & $0.92$ & $0.823$ \\
     \hline
\end{tabular}
\end{center}
\end{table}
 
\begin{table}[!h]
    \caption{Comparison with existing results on AVA dataset}
    \label{tab:table5}

    \begin{center}
    \begin{tabular}{|c|c|c|}
    \hline
         Network & Accuracy \\
         \hline
         Signle Column Network (SCNN) \cite{six} & 71.20 \\
         \hline
         Double Column Network (DCNN) \cite{six} & 73.25 \\
         \hline
         Brain Inspired Deep Neural Network (BDN) \cite{seven} & 78.08 \\
         \hline
         \textbf{Triple Column Network} & \textbf{82.3} \\
         \hline
    \end{tabular}
    \end{center}
\end{table}

\section{Conclusions} 
After conducting various results on AVA dataset for different architectures and observing the results, we came to a conclusion that increasing the number of columns in the architecture did give us better results compared to single column architecture results. We also observed that compared to other proposed architectures like SCNN \cite{six}, DCNN \cite{six} and BDN \cite{seven}, triple column architecture showed the best results that is $82.3\%$ accuracy was achieved. The major reason for achieving such a high accuracy was due to the fact of involving different forms of images such as cropped, padded and saliency maps. Giving balanced and equal weights to these forms of an image helped to train the network more efficiently. 
\label{sec:con}

\end{document}